\documentclass[a4paper]{article}

\usepackage{INTERSPEECH2021}
\usepackage{times}
\usepackage{booktabs}
\usepackage[ruled,longend]{algorithm2e}
\usepackage{amsmath, amssymb}
\usepackage{mathtools}
\usepackage{latexsym}
\usepackage{todonotes}
\usepackage{hyperref}
\usepackage{comment}
\usepackage{multirow}
\usepackage{tabto}

\usepackage{float} 
\usepackage{siunitx}
\usepackage{cleveref}

\newcommand{\cf}{c.\,f.\xspace}

\newcommand{\nameproject}{\textit{LeBenchmark}}

    \title{{\nameproject}: A Reproducible Framework for Assessing \\ 
    Self-Supervised Representation Learning from Speech
}
\name{Solène Evain$^{1,*}$, Ha Nguyen$^{1,2,*}$, Hang Le$^{1,*}$, Marcely Zanon Boito$^{1,*}$, Salima Mdhaffar$^{2,*}$, Sina Alisamir$^{1,3}$, Ziyi Tong$^1$, Natalia Tomashenko$^2$, Marco Dinarelli$^{1,*}$, Titouan Parcollet$^{2,*}$, Alexandre Allauzen$^4$, Yannick Estève$^2$, Benjamin Lecouteux$^1$, François Portet$^1$, Solange Rossato$^1$, Fabien Ringeval$^1$,  Didier Schwab$^1$ and Laurent Besacier$^{1,5}$}
\address{\normalsize 
  $^1$Univ. Grenoble Alpes, CNRS, Inria, Grenoble INP, LIG, 38000 Grenoble, France  \qquad
  $^2$LIA, Avignon Universit{\'e}, France \\
  $^3$Atos, {\'E}chirolles, France \qquad
  $^4$ESPCI, CNRS LAMSADE, PSL Research University, France \qquad
  $^5$Naver Labs Europe, France \\
  $^*$Equal contributors}
\email{
yannick.esteve@univ-avignon.fr, francois.portet@univ-grenoble-alpes.fr}
\begin{document}

\maketitle
\begin{abstract}
  Self-Supervised Learning~(SSL) using huge unlabeled data has been successfully explored for image and natural language processing. Recent works also investigated SSL from speech. 
  They were notably successful to improve performance on downstream tasks such as automatic speech recognition (ASR). While these works suggest it is possible to reduce dependence on labeled data for building efficient speech systems, their evaluation was mostly made on ASR and using multiple and heterogeneous experimental settings (most of them for English). This questions the objective comparison of SSL approaches and the evaluation of their  impact on building speech systems.
  In this paper, we propose {\nameproject}: a reproducible framework for assessing SSL from speech. 
  It not only includes ASR (high and low resource) tasks but also spoken language understanding, speech translation and emotion recognition. We also focus on speech technologies in a language different than English: French. 
  SSL models of different sizes are trained from carefully sourced and documented datasets.
  Experiments show that SSL is beneficial for most but not all tasks which confirms the need for exhaustive and reliable benchmarks to evaluate its real impact. {\nameproject} is shared with the scientific community for reproducible research in SSL from speech.
  \end{abstract}

\section{Introduction}

Self-Supervised Learning~(SSL) based on huge unlabeled data has been explored  successfuly for image processing
~\cite{bachman2019amdim,chen2020simple} and Natural Language Processing (NLP)~\cite{bert}. 
Recently, 
pioneering work investigated SSL from speech, and successfully improved
performance on downstream tasks such as speech recognition in low-resource scenarios~\cite{Baevski, kawakami2020learning}. 
One observation that can be made about those recent studies on SSL for speech is that, as common benchmarks are not experimented, comparison of different SSL approaches are difficult to make.
In addition, contributions have mostly been done on English, with a few recent studies related to multilingual SSL~\cite{conneau2020unsupervised,wang2021voxpopuli}.
We propose to remedy these shortcomings by providing a reproducible benchmark\footnote{\url{https://github.com/LeBenchmark/}} that includes:
\begin{itemize}
   
    \item a large and heterogeneous collection of French speech utterances (read, prepared, and spontaneous);
    
   
    \item pre-trained SSL models learnt on collections of 1k and 3k hours of French speech;
    
    \item assessments on Speech Recognition (ASR), Spoken Language Understanding (SLU), Speech Translation (AST) and  Emotion Recognition (AER) in French.
\end{itemize}

\section{Background}

\begin{table*}
\centering
\caption{Statistics for the speech corpora used to train SSL models according to gender information (male / female / unknown). The small dataset (1k hours) is from MLS only, and the medium dataset (2.9k hours) is from all of them; duration: hour(s):minute(s).
}
\resizebox{15cm}{!}{
\scriptsize

\begin{tabular}{lccccc}\toprule
\textbf{Corpus}                                                    & \textbf{\# Utterances}               & \textbf{Duration}                    & \textbf{\# Speakers} & \textbf{Mean Utt. Duration} & \textbf{Speech type}\\
\midrule
\begin{tabular}[l]{@{}l@{}}African Accented \\ French~\cite{noauthor_african_2003}\end{tabular} & \begin{tabular}[c]{@{}c@{}}\textbf{16,402} \\ 373 / 102 / 15,927\end{tabular}                      & \begin{tabular}[c]{@{}c@{}}\textbf{18:56} \\ -- / -- / 18:56\end{tabular}                      & \begin{tabular}[c]{@{}c@{}}\textbf{232} \\ 48 / 36 / 148\end{tabular}         & 
\begin{tabular}[c]{@{}c@{}}\textbf{4\,s} \\ -- / -- / --\end{tabular}                        &  Read\\
\hline
Att-Hack~\cite{le_moine_att-hack_2020}                                                           & \begin{tabular}[c]{@{}c@{}}\textbf{36,339} \\ 16,564 / 19,775 / 0\end{tabular}                     & \begin{tabular}[c]{@{}c@{}}\textbf{27:02} \\ 12:07 / 14:54 / 0:00\end{tabular}                       & \begin{tabular}[c]{@{}c@{}}\textbf{20} \\ 9 / 11 / 0\end{tabular}                  & \begin{tabular}[c]{@{}c@{}}\textbf{2.7\,s}\\ 2.6\,s / 2.7\,s / --\end{tabular}                        & \begin{tabular}[c]{@{}c@{}}Acted \\ Emotional\end{tabular}          \\\hline
CaFE~\cite{gournay_canadian_2018}                                                               & \begin{tabular}[c]{@{}c@{}}\textbf{936} \\ 468 / 468 / 0\end{tabular}                              & \begin{tabular}[c]{@{}c@{}}\textbf{1:09} \\ 0:32 / 0:36 / 0:00\end{tabular}                          & \begin{tabular}[c]{@{}c@{}}\textbf{12} \\ 6 / 6 / 0\end{tabular}                   & \begin{tabular}[c]{@{}c@{}}\textbf{4.4\,s} \\ 4.2\,s / 4.7\,s / --\end{tabular}                        & \begin{tabular}[c]{@{}c@{}}Acted \\ Emotional\end{tabular}          \\\hline
CFPP2000*~\cite{branca-rosoff_discours_2012} \cite{cfpp2000_website}                                                           & \begin{tabular}[c]{@{}c@{}}\textbf{9853} \\ 166 / 1,184 / 8,503\end{tabular}                    & \begin{tabular}[c]{@{}c@{}}\textbf{16:26}\\ 0:14 / 1:56 / 14:16\end{tabular}                      & \begin{tabular}[c]{@{}c@{}}\textbf{49} \\ 2 / 4 / 43\end{tabular}                  & \begin{tabular}[c]{@{}c@{}}\textbf{6.0\,s}\\ 5.0\,s / 5.0\,s / 6.0\,s\end{tabular}                        & Spontaneous                                                          \\\hline
ESLO2~\cite{eshkol-taravella_grand_2011}, \cite{eslo_website}                                                              & \begin{tabular}[c]{@{}c@{}}\textbf{62,918} \\ 30,440 / 32,147 / 331\end{tabular}                          & \begin{tabular}[c]{@{}c@{}}\textbf{34:12} \\ 17:06 / 16:57 / 0:09 \end{tabular}                          & \begin{tabular}[c]{@{}c@{}}\textbf{190} \\ 68 / 120 / 2\end{tabular}                & \begin{tabular}[c]{@{}c@{}}\textbf{1.9\,s} \\ 2.0\,s / 1.9\,s / 1.7\,s\end{tabular}                        & Spontaneous                                                          \\\hline
EPAC**~\cite{esteve_epac_2010}                                                             & \begin{tabular}[c]{@{}c@{}}\textbf{623,250} \\ 465,859 / 157,391 / 0\end{tabular}                  & \begin{tabular}[c]{@{}c@{}}\textbf{1,626:02} \\ 1,240:10 / 385:52 / 0:00\end{tabular}                & \begin{tabular}[c]{@{}c@{}}\textbf{1,935} \\ -- / -- / --\end{tabular}           & \begin{tabular}[c]{@{}c@{}}\textbf{9\,s} \\ -- / -- / --\end{tabular}                       & \begin{tabular}[c]{@{}c@{}}Radio \\ Broadcasts\end{tabular}          \\\hline
GEMEP~\cite{banziger_introducing_2012}                                                              & \begin{tabular}[c]{@{}c@{}}\textbf{1,236} \\ 616 / 620 / 0\end{tabular}                            & \begin{tabular}[c]{@{}c@{}}\textbf{0:50} \\ 0:24 / 0:26 / 0:00\end{tabular}                          & \begin{tabular}[c]{@{}c@{}} \textbf{10} \\ 5 / 5 / 0\end{tabular}                                                              & \begin{tabular}[c]{@{}c@{}}\textbf{2.5\,s} \\ 2.4\,s / 2.5\,s / --\end{tabular}                        & \begin{tabular}[c]{@{}c@{}}Acted \\ Emotional\end{tabular}          \\\hline
MLS French~\cite{pratap_mls_2020}                                                      & \begin{tabular}[c]{@{}c@{}}\textbf{263055} \\ 124,590 / 138,465 / 0\end{tabular}                   & \begin{tabular}[c]{@{}c@{}}\textbf{1,096:43} \\ 520:13 / 576:29 / 0:00\end{tabular}                  & \begin{tabular}[c]{@{}c@{}}\textbf{178} \\ 80 / 98 / 0\end{tabular}                & \begin{tabular}[c]{@{}c@{}}\textbf{15.0\,s} \\ 15.0\,s / 15.0\,s / --\end{tabular}                     & Read                                                                 \\\hline
MPF~\cite{MPF_paper_2017}, \cite{MPF_ortolang}                                                                & \begin{tabular}[c]{@{}c@{}}\textbf{19,527} \\ 5,326 / 4,649 / 9,552\end{tabular}                          & \begin{tabular}[c]{@{}c@{}}\textbf{19:06} \\ 5:26 / 4:36 / 9:03\end{tabular}                            & \begin{tabular}[c]{@{}c@{}}\textbf{114}\\ 36 / 29 / 49\end{tabular}                 & \begin{tabular}[c]{@{}c@{}}\textbf{3.5\,s} \\ 3.7\,s / 3.6\,s / 3.4\,s\end{tabular}                        & Spontaneous                                                          \\\hline
PORTMEDIA (French)~\cite{lefevre_robustesse_2012}                                                          & \begin{tabular}[c]{@{}c@{}}\textbf{19,627} \\ 9,294 / 10,333 / 0\end{tabular}                      & \begin{tabular}[c]{@{}c@{}}\textbf{38:59}\\ 19:08 / 19:50 / 0:00\end{tabular}                        & \begin{tabular}[c]{@{}c@{}}\textbf{193} \\ 84 / 109 / 0\end{tabular}               & \begin{tabular}[c]{@{}c@{}}\textbf{7.1\,s} \\ 7.4\,s / 6.9\,s / --\end{tabular}                        & \begin{tabular}[c]{@{}c@{}}Acted telephone \\ dialogue \end{tabular} \\\hline
TCOF (Adult\,s)~\cite{TCOF_ortolang}                                                              & \begin{tabular}[c]{@{}c@{}}\textbf{58,722} \\ 10,377 / 14,763 / 33,582\end{tabular}                & \begin{tabular}[c]{@{}c@{}}\textbf{53:59} \\ 9:33 / 12:39 / 31:46\end{tabular}                     & \begin{tabular}[c]{@{}c@{}}\textbf{749} \\ 119 / 162 / 468\end{tabular}            & \begin{tabular}[c]{@{}c@{}}\textbf{3.3\,s} \\ 3.3\,s / 3.1\,s / 3.4\,s\end{tabular}                        & Spontaneous                                                          \\\midrule
\textbf{ALL}                                                       & 
\begin{tabular}[c]{@{}c@{}}\textbf{1,111,865} \\ 664,073 / 379,897
 / 67,895\end{tabular} & \begin{tabular}[c]{@{}c@{}}\textbf{2,933:18} \\ 1,824:53 / 1034:15 / 74:10\end{tabular} & -                                                                       & -                                                                            & -  
                                                    \\\bottomrule   
\end{tabular} }
 \\\small *version without the CEFC corpus v2.1, 02/2021; **speakers are not uniquely identified.\\
    \label{tab:medium_and_small_datasets}
\end{table*}

Most deep learning methods highly rely on large quantities of labeled training data. Particularly, current acoustic models require thousands of hours of transcribed speech to achieve state-of-the-art performance. 
However, this requirement 
cannot be fullfiled by the majority of the nearly 7,000 languages spoken worldwide.
To overcome this, SSL has been recently proposed as an interesting alternative for data representation learning, as it requires 
less or no annotated data. Such learnt representations have been very successful in  vision~\cite{bachman2019amdim,chen2020simple}
and NLP~\cite{bert, peters2018deep}. 
Self-supervised learning from speech consists of resolving \textit{pseudo-tasks}, which do not require human annotation, as a pre-training for the real tasks to solve. These \textit{pseudo-tasks} target predicting the next samples, or solving ordering problems. For instance, Autoregressive Predictive Coding~(APC) considers the sequential structure of speech and predicts information about a future frame~\cite{DBLP:journals/corr/abs-1904-03240, chung2020improved}, whereas Contrastive Predictive Coding~(CPC) 
distinguishes a future speech frame from distractor samples~\cite{Baevski, Schneider2019}, which is an easier learning objective compared to APC. 
Such representations have been shown to improve performance in several speech tasks~\cite{chung2020generative}, while being less sensitive to domain and/or language mismatch~\cite{kawakami2020learning}.
It has also been shown that features extracted through a CPC pre-training can be transfered to other languages, with performance being on par or superior to a supervised pre-training~\cite{riviere2020unsupervised}.



\section{Gathering a Large and Heterogeneous Speech Collection in French}
\label{sec:data}

Recently, large multilingual corpora that include French have been made available, such as MLS~\cite{pratap_mls_2020}~(1,096\,h), or voxpopuli~\cite{wang2021voxpopuli}~(+4,500\,h). However, these are restricted to either read or well-prepared speech, failing to provide diversity in the speech samples, such as accented, spontaneous and/or affective speech. As a consequence, SSL models trained only on these corpora may present poor generalisation abilities on spontaneous or affective speech.
In this work, 
we gathered a large variety of speech corpora in French that cover different accents~(MLS, African Accented Speech, CaFE), acted emotions~(GEMEP, CaFE, Att-Hack), telephone dialogues~(PORTMEDIA), read~(MLS, African Accented French) and spontaneous sentences~(CFPP2000, ESLO2, MPF, TCOF), as well as broadcast speech~(EPAC). 
Compared to MLS and Voxpopuli, 
our dataset is more diverse, carefully sourced and contains detailed metadata (speech type, and speaker gender), which would facilitate future fine-grained analysis of SSL such as training gender/style specific models.
Moreover, 
our dataset has a more realistic representation of speech turns in real life, compared to MLS~(see average utterance duration per speaker in Table~\ref{tab:medium_and_small_datasets}).
We detail below the necessary steps for producing the dataset 
whose statistics are reported in Table~\ref{tab:medium_and_small_datasets}.

\noindent\textbf{Pre-processing for SSL training} 
Audio recordings were segmented using time stamps from speech transcriptions. We also retrieved, when available, speaker labels and gender information. Following ~\cite{baevski2020wav2vec}, we removed utterances shorter than 1\,s, and longer than 30\,s. Finally, when necessary, audio segments were converted to mono PCM files using 16\,bits and a sampling frequency of 16\,kHz.

\noindent\textbf{Final dataset} To the best of our knowledge, this is the very first study that explores such a diverse and large ensemble of datasets for SSL training. 
It includes 2,933\,h of speech, from which 1,115\,h is read speech, 1,626\,h broadcast speech, 124\,h spontaneous speech, 38\,h acted telephone dialogues, and 29\,h acted emotional speech. Regarding gender, we collected 1,824\,h of speech from male speakers, 1,034\,h from female speakers, and 74\,h from unknown gender. 
This full corpus is refered as \textit{medium}, and the subset made of MLS only is refered as \textit{small}.

\section{Training SSL Models}
\label{sec:ssl_trained}

\textit{LeBenchmark} provides four Wav2Vec2.0 models~\cite{baevski2020wav2vec} pretrained on the gathered French data described in Section \ref{sec:data}. Following \cite{baevski2020wav2vec}, two different Wav2Vec2.0 architectures (\textit{large} and \textit{base}) are coupled with our \textit{small} (S) and \textit{medium} (M) corpus to form our set of Wav2Vec2.0 models: W2V2-Fr-S-\textit{base}, W2V2-Fr-S-\textit{large}, W2V2-Fr-M-\textit{base}, W2V2-Fr-M-\textit{large}.
Hyperparameters and architectures for \textit{base} and \textit{large} are identical to the ones first introduced in \cite{baevski2020wav2vec}. All models are trained on four Nvidia Tesla V100 (32GB) until the loss observed on the validation set of the MLS corpus (Section \ref{sec:data}) reaches a stable point. Pretrained Wav2Vec2.0 models are shared with the community via HuggingFace\footnote{\url{https://huggingface.co/LeBenchmark}} for further integration with well-known toolkits such as Fairseq~\cite{ott2019fairseq}, SpeechBrain~\cite{SB2021} or Kaldi~\cite{povey2011kaldi}. In some downstream experiments mentioned below, we also use two Wav2Vec2.0 (\textit{base} / \textit{large}, no finetuning) English models pre-trained on the full LibriSpeech ($960$\,h) corpus by Fairseq,\footnote{\url{https://github.com/pytorch/fairseq/tree/master/examples/wav2vec}} and refer to them as \textit{W2V2-En base} and \textit{W2V2-En large}. The XLSR-53-\textit{large}~\cite{conneau2020unsupervised} multilingual model is also used.

\section{Benchmarking our SSL Models }

\subsection{Automatic Speech Recognition}
We evaluate the contribution of SSL for 
ASR using a hybrid DNN-HMM and an end-to-end approach.

\noindent\textbf{Datasets} The ASR tasks target two different types of corpora: Common Voice~\cite{ardila2020common} and ETAPE~\cite{gravier2012etape}. 
Common Voice is a very large crowdsourced corpus (477\,h) of read speech in French with transcripts -- training: 428\,h, development: 24\,h, and test: 25\,h,
while ETAPE is a smaller (36\,h) but more challenging corpus composed of diverse French TV broadcast programs -- training: 22\,h, development: 7\,h, and test: 7\,h.

\noindent\textbf{Hybrid DNN-HMM} 
 The baseline acoustic models~(AM) have been trained on 40-dimensional high-resolution (\textit{hires}) MFCC features using the Kaldi~\cite{povey2011kaldi} toolkit with a state-of-the-art factorized time delay neural network~(TDNN-F)~\cite{povey2018semi,peddinti2015time} 
 on the ETAPE training corpus~\cite{gravier2012etape} only. 
 The model has 12 TDNN-F layers (1,024-dimensional, with projection dimension of 128) and a 3,432-dimensional output layer. 
 It was trained  using lattice-free maximum mutual information (LF-MMI)~\cite{povey2016purely}  and cross-entropy criteria.
 Speed and volume perturbations have been applied for data augmentation. 
We used a similar topology to train  three other systems with different types of input features extracted by W2V2-Fr-M-\textit{large}, W2V2-En-\textit{large}~\cite{baevski2020wav2vec}, and  XLSR-53-\textit{large} models.
100-dimensional speaker i-vectors were appended to the input features for all the models.
Two trigram LMs were used in evaluation: 
(1) a larger one with a 82k vocabulary
and (2) a smaller one trained on ETAPE training data only with a 17.5k vocabulary.

\noindent\textbf{End-to-End} 
Our end-to-end system is defined with the SpeechBrain toolkit~\cite{SB2021} using  
an encoder/decoder architecture with attention: 
the encoder is 
a Convolutionnal Recurrent Deep Neural Network CRDNN (VGG + RNN + DNN),
and the decoder is a joint CTC/Attention LSTM neural network.
When used with the Wav2Vec2.0 features (same from 
hybrid DNN-HMM ASR experiments),
the CNN blocks are removed from the CDRNN encoder.
For end-to-end ASR experiments, 
the neural network output corresponds to 500 byte pair encoding (BPE) units~\cite{sennrich2016neural} computed on the manual transcriptions of the respective training datasets.
No additional language model is used in these experiments, neither data augmentation.
For comparison purposes, we also use 80-dimension log Mel filterbank~(MFB) features.

\noindent \textbf{Results} 
The WER results on the ETAPE development and test data sets for the hybrid DNN-HMM models are given in Table~\ref{tab:res_kaldi_models_lm_etape}. 
Among the  models trained on SSL features, two models provide improvement over the baseline AM trained on MFCC features: the model
trained on XLSR-53 features (7--8\% of relative WER reduction) and the model
trained on W2V2-Fr-M-\textit{large} features (17--20\% of relative WER reduction). To our knowledge, this is the first time SSL features are used for hybrid DNN-HMM ASR. Actually, the hybrid DNN-HMM ASR system is much better than its end-to-end counterpart on ETAPE (see next paragraph).
This is partly due to the
use of speaker adaptation (i-vectors) and hand-crafted pronunciation dictionary which might be particularly beneficial to the hybrid DNN-HMM system, compared to end-to-end ASR, for the low resource ETAPE task.

Table~\ref{tab:res_cv_sb_models} presents the results achieved with end-to-end ASR on Common Voice 6.1 and 
on ETAPE datasets. On the ETAPE, filterbank parameters (MFB) got significanly the best results, while on Common Voice, W2V2-Fr-M-\textit{large} is very close. In all  (hybrid and end-to-end) ASR experiments,  among the wav2vec models, W2V2-Fr-M-\textit{large} got the best results.


    

\begin{table}
\caption{ASR results (WER,\%) on the ETAPE corpus
for hybrid DNN-HMM acoustic models with TDNN-F topology.}
  \footnotesize
  \label{tab:res_kaldi_models_lm_etape}
  \centering
  \begin{tabular}{l|c|c||c|c}
    \toprule
    \textbf{Language Model}	 &	\multicolumn{2}{c||}{\textbf{\scriptsize{ETAPE}}} & \multicolumn{2}{c}{\textbf{\scriptsize{ESTER-1.2 + EPAC}}}	\\  \midrule
    \textbf{Features}	 &	\textbf{Dev} &	\textbf{Test} &	\textbf{Dev} &	\textbf{Test}\\  \midrule
  hires MFCC	&	39.28 &	40.89 &	35.60 &	37.73 \\ \hline
W2V2-Fr-M-\textit{large}   & \textbf{32.19} & 	\textbf{33.87} & \textbf{28.53} & \textbf{30.77}\\
W2V2-En-\textit{large} &	39.93  & 	42.30 &	36.18 & 	38.75 \\ 
XLSR-53-\textit{large} &	36.36 &	38.19 &	32.81 &	35.17 \\
    \bottomrule
  \end{tabular}
\end{table}
\normalsize


\begin{table}
  \caption{End-to-end ASR results (WER,\%) on Common Voice and ETAPE corpora. 
  $(*)$~means the training algorithm did not converge to a WER smaller than 100\%.}
  \footnotesize
  \label{tab:res_cv_sb_models}
  \centering
  \begin{tabular}{l|c|c||c|c}
    \toprule
        {\textbf{Corpus}} & \multicolumn{2}{c||}{\textbf{CommonVoice}} & \multicolumn{2}{c}{\textbf{ETAPE}} 
        \\ \midrule
    \textbf{Features}	 &	\textbf{Dev} &	\textbf{Test} & \textbf{Dev} & \textbf{Test}\\  \midrule
  MFB &	\textbf{20.19}&	\textbf{23.40} & \textbf{54.55} & \textbf{56.17}\\ \hline
W2V2-Fr-M-\textit{large} &20.23& 24.06 & 55.56& 57.04\\ 
W2V2-En-\textit{large} &34.07&37.29 & 98.79& 99.10\\ 
XLSR-53-\textit{large} &30.07 &32.72 & $(*)$ & $(*)$\\
    \bottomrule
  \end{tabular}
\end{table}
\normalsize


\subsection{Spoken Language Understanding}


\noindent\textbf{Dataset} 
The MEDIA corpus~\cite{Bonneau-Maynard2006-media,dinarelli-icassp-2020} is used for the French SLU benchmark. 
The corpus is made up of 12,908 utterances (41.5\,h) for training, 1,259 utterances (3.5\,h) for development and 3,005 utterances (11.3\,h) for test.

\noindent\textbf{Model} Our end-to-end model has a pyramidal LSTM encoder similar to \cite{7472621}. The decoder integrates, in addition to the usual attention mechanism for attending the encoder hidden states, an attention mechanism for attending all previous decoder prediction's embeddings, instead of just the previous one \cite{DBLP-journals-corr-BahdanauCB14}.
We use an incremental training strategy similar to \cite{dinarelli-icassp-2020}, by first training an ASR model from scratch which is used to initialize parameters of a SLU model using a simple linear layer as decoder; and then using this simple SLU 
to initialize parameters of our final SLU model, which uses a LSTM decoder.
The model, which is implemented using \textit{Fairseq}~\cite{ott2019fairseq}, has the same settings 
as~\cite{dinarelli-icassp-2020} to allow direct and fair comparison.



\begin{table}
\caption{End-to-end SLU results on the MEDIA corpus.}
\begin{center}
\resizebox{7.7cm}{!}{
    \begin{tabular}{l|l|c|c}
    \toprule
        \textbf{Model} & \textbf{Features} & \textbf{Dev} & \textbf{Test} \\
        \midrule
        \multicolumn{4}{c}{\textbf{Token decoding (Word Error Rate \%)}} \\
        \hline
        \cite{dinarelli-icassp-2020} Seq    &   spectrogram &   \textbf{29.42}   &   \textbf{28.71} \\
        \hline
        Kheops$\oplus$Basic &   spectrogram &   36.25  &   37.12 \\
        Kheops$\oplus$LSTM &   spectrogram &   \textbf{35.37}  &   \textbf{35.98} \\
        \hline
        Kheops$\oplus$Basic &   W2V2-En-\textit{base} &   19.80  &   21.78 \\
        Kheops$\oplus$Basic &   W2V2-En-\textit{large} &   24.44  &   26.96 \\
        \hline
        Kheops$\oplus$Basic &   W2V2-Fr-S-\textit{base} &   23.11  &   25.22 \\
        Kheops$\oplus$Basic &   W2V2-Fr-S \textit{large} &   18.48  &   19.92 \\
        Kheops$\oplus$Basic &   W2V2-Fr-M-\textit{base} &   14.97 &   16.37 \\
        Kheops$\oplus$Basic &   W2V2-Fr-M \textit{large} &   \textbf{11.77} &   \textbf{12.85} \\ 
        \hline
        Kheops$\oplus$Basic &   XLSR-53-\textit{large} &   14.98 &   15.74 \\
        \toprule
        \multicolumn{4}{c}{  \textbf{SLU decoding (Concept Error Rate \%)}} \\
        \midrule
        \cite{dinarelli-icassp-2020} Seq    &   spectrogram &   28.11   &   27.52 \\
        \cite{dinarelli-icassp-2020} XT    &   spectrogram &   \textbf{23.39}   &   \textbf{24.02} \\
        \hline
        Kheops$\oplus$Basic           &   spectrogram &   39.66  &   40.76 \\
        Kheops$\oplus$Basic +token    &   spectrogram &   34.38  &   34.74 \\
        Kheops$\oplus$LSTM +SLU    &   spectrogram &    \textbf{33.63}  &   \textbf{34.76} \\
        \hline
        Kheops$\oplus$LSTM           &   W2V2-En-\textit{base} &   26.31  &   26.11 \\
        Kheops$\oplus$LSTM           &   W2V2-En-\textit{large} &   28.38  &   28.57 \\
        \hline
        Kheops$\oplus$LSTM           &   W2V2-Fr-S-\textit{base} &   26.16  &   26.69 \\
        Kheops$\oplus$LSTM           &   W2V2-Fr-S \textit{large} &   22.53  &   23.03 \\
        Kheops$\oplus$LSTM         &   W2V2-Fr-M-\textit{base} &   22.56  &   22.24 \\
        Kheops$\oplus$LSTM           &   W2V2-Fr-M-\textit{large} &   \textbf{18.54}  &   \textbf{18.62} \\
        \hline
        Kheops$\oplus$LSTM           &   XLSR-53-\textit{large} &   20.34  &   19.73 \\
        \bottomrule
    \end{tabular}}
\end{center}
\label{tab:End2EndSLUResults}
\end{table}

\noindent\textbf{Results} for ASR and SLU obtained with different speech representations are shown in Table~\ref{tab:End2EndSLUResults}, and they are given in terms of Word Error Rate (WER) and Concept Error Rate (CER) respectively, which is computed the same way as WER but on concept sequences.
The ASR results are included because 
we use token-level models~(ASR) to pre-initialize SLU models. 
The $\oplus$ symbol is used for separating Encoder and Decoder names: \emph{Kheops} is the pyramidal encoder inspired from~\cite{7472621}, \emph{Basic} is the linear decoder, and \emph{LSTM} is the more advanced LSTM decoder. 
For ASR, using SSL features as input resulted in an impressive drop in WER, 
even when using English SSL models.
At best, we achieve a WER of $11.77$\% on the development data with the W2V2-Fr-M-\textit{large} features.
SLU results (\textit{SLU decoding} in Table~\ref{tab:End2EndSLUResults}) follow the same trend. 
The best performance is obtained again with W2V2-Fr-M-\textit{large} features, with a CER of $18.54$ on the development data. This result improves 
previous work by almost $5$ points ($23.39$~vs.~$18.54$), and stands as the new state-of-the-art result using only MEDIA training data for learning SLU models. Better results have been obtained in~\cite{Caubriere2019,tomashenko2019recent} by using more transcribed and annotated data, 
in addition to the MEDIA corpus, via transfer learning.

\subsection{Speech-to-text Translation}
Automatic speech-to-text translation (AST) consists in translating a speech utterance in a source language to a text in a target language. In this work, we are interested in translating from French to another language.

\noindent\textbf{Datasets} We selected subsets having French as the source language in two large multilingual speech corpora: CoVoST-2~\cite{wang2020covost2} and multilingual TEDx~\cite{salesky2021multilingual}.
Our benchmark covers translation directions from French to three 
target languages, 
English (\texttt{en}), Portugese (\texttt{pt}), and Spanish (\texttt{es}), with following training sizes: 50\,h~(TEDx/\texttt{en}), 38\,h~(TEDx/\texttt{es}), 25\,h~(TEDx/\texttt{pt}), and 180\,h~(CoVoST2/\texttt{en}). 

\noindent\textbf{Features} We compared models using 
80-dimensional MFB features and SSL representations. In addition to the four French Wav2Vec2.0 models trained in Section~\ref{sec:ssl_trained}, we also considered the following off-the-shelf models: English~\cite{baevski2020wav2vec} (W2V2-En-\textit{base/large}), French~\cite{wang2021voxpopuli} (W2V2-Fr-VP-\textit{base/large}), and the multilingual model XLSR-53~\cite{conneau2020unsupervised} (XLSR-53-\textit{large}). For a fair comparison, we did not use additional data augmentation technique nor ASR encoder pre-training in the experiments.

\noindent\textbf{Models} We trained Transformer~\cite{vaswani2017attention} models using the \textsc{fairseq s2t} toolkit~\cite{wang2020fairseq}, 
and using a small architecture with 12-layers encoder, 6-layers decoder, and hidden dimension $D=256$. 
For models using SSL features, we inserted a block of Linear-ReLU before convolutional layers not only to reduce the number of parameters~\cite{nguyen2020investigating}, but also because we preliminary observed improved performance with this block.

\begin{table}[!t]
\caption{AST results (BLEU) on dev/valid and test sets of CoVoST-2 (CV2) and multilingual TEDx (mTEDx).} 

\centering
\resizebox{\linewidth}{!}{

\begin{tabular}{l|c | ccc | c | ccc }
	\toprule
	\multirow{3}{*}{\textbf{Input features}}& \multicolumn{4}{c|}{\textbf{Dev/Valid data}} & \multicolumn{4}{c}{\textbf{Test data}} \\
	\cmidrule{2-9}
	 & \textbf{CV2} & \multicolumn{3}{c|}{\textbf{mTEDx}} & \textbf{CV2} & \multicolumn{3}{c}{\textbf{mTEDx}}\\
	&\textbf{en} & \textbf{en} & \textbf{es} & \textbf{pt}  &\textbf{en} & \textbf{en} & \textbf{es} & \textbf{pt} \\
	\midrule
	MFB & \textbf{23.37} & 1.14 & 0.84 & 0.49 & 
	
	\textbf{22.66} & 1.33 & 0.98 & 0.68\\
	\midrule
	W2V2-En-\textit{base} & 19.24 & 0.90 & 0.65 & 0.43 & 
	18.19 & 0.88 & 0.34 & 0.27 \\
	
	W2V2-En-\textit{large} & 17.07 & 0.75 & 0.61 & 0.45 & 
	16.45 & 0.85 & 0.67 & 0.32\\
	
	\midrule
	W2V2-Fr-S-\textit{base} & 19.86 & 2.64 & 0.49 & 0.50 &
	19.04 &1.66& 0.67& 0.61\\
	
	W2V2-Fr-S-\textit{large} & 19.62 & 5.12 & 4.62 & \textbf{2.06} & 
	18.61 & 2.97&3.19 &\textbf{2.25}\\
	
	\midrule
	W2V2-Fr-M-\textit{base} & 19.47 & 6.98& 1.87& 0.63& 
	18.32 & 6.37 & 1.99 & 0.54\\
	
	W2V2-Fr-M-\textit{large} & {20.17} & \textbf{9.35} & \textbf{7.72}& {1.58}& 
	19.35 & \textbf{6.76}& \textbf{6.63}& {1.63}\\
	
	\midrule
	W2V2-Fr-VP-\textit{base} & 18.44 & 0.81& 0.45& 0.56& 
	17.40 & 0.89& 0.58& 0.75\\
	
	W2V2-Fr-VP-\textit{large} & 20.72 & 7.43 & 4.66 & 0.43& 
	19.88 & 5.39& 3.62& 0.49\\
	\midrule
	XLSR-53-\textit{large} & 20.54 & 0.59& 0.41& 0.49& 
	19.93 & 0.44& 0.62&0.29 \\
	
	\bottomrule
\end{tabular}
}
\label{tab:ast_results_transformer}
\end{table}

    

\noindent\textbf{Results} 
shown in Table~\ref{tab:ast_results_transformer} highlight the benefit of SSL features only in medium and low-resource settings, namely mTEDx:  
our W2V2-Fr-M-\textit{large} produces the best results across all language pairs, except for \texttt{pt} which is too low-resourced to obtain decent BLEU whatever the features used.
In the higher-resource scenario (CV2), however, the best-performing SSL features are still $2.65$ BLEU point below the MFB ones. 



\subsection{Automatic Emotion Recognition}
Automatic emotion recognition aims at detecting human's apparent emotions from sensors such as microphones and cameras. Affective computing has many useful applications in the domain of health, education, art and entertainment.

\noindent\textbf{Datasets} 
We used the RECOLA dataset \cite{ringeval2013introducing}, which contains 3.8\,h of noise-free recordings of spontaneous interactions between French-speaking subjects 
solving a collaborative task in remote condition -- training, development and test partitions include each one third of the data, and AlloSat \cite{macary2020allosat}, a more recent corpus containing 37\,h of real-life call center conversations in French -- training: 25.6\,h, development: 5.8\,h, and test: 6.0\,h. Both datasets are annotated by several annotators using time-continuous dimensions which are averaged to define an emotion \textit{gold-standard}: arousal (from passive to active) and valence (from negative to positive) for RECOLA, and a dimensional axis ranging from frustration to satisfaction for AlloSat.

\noindent\textbf{Features}
We extracted 40-dimensional MFB features 
that were standardized to zero mean and unit standard deviation
according to the training set, and SSL features that were pre-processed by a normalisation layer. Annotations were resampled to match the sampling frequency of the features, which was 100\,Hz for MFB and 50\,Hz for the Wav2Vec models. 

\noindent\textbf{Models}
We used 
a simple model based on a linear layer mapping features to one emotional dimension, followed by a tangent hyperbolic function (Linear-Tanh). The other model is a 1-layer GRU with the hidden layer $D=[32, 64]$, followed by the Linear-Tanh function. Adam optimiser 
was used 
and patience was set to 
15\,epochs, and the Concordance Correlation Coefficient~\cite{LI89-ACC} was used as loss function to train the models as in~\cite{Weninger16-DTR,Trigeorgis16-AFE}.


\noindent\textbf{Results}
\begin{table}
\caption{AER results (Concordance Correlation Coefficient of emotion predictions) on the RECOLA and AlloSat test sets.}
\scriptsize
\begin{center}
\resizebox{8cm}{!}{
    \begin{tabular}{l|c|c|c|c}
        \toprule
        \multicolumn{2}{c|}{\textbf{Corpus}} & \multicolumn{2}{c|}{\textbf{RECOLA}} & \textbf{AlloSat} \\ \midrule
        \textbf{Model} & \textbf{Feature} & \textbf{Arousal} & \textbf{Valence} & \textbf{Satisfaction} \\ \midrule
        Linear-Tanh & MFB & 0.192 & 0.075 & 0.065 \\
        Linear-Tanh & W2V2-Fr-M-\textit{base} & \textbf{0.385} & \textbf{0.090} & \textbf{0.193} \\
        Linear-Tanh & XLSR-53-\textit{large} & 0.155 & 0.024 & 0.093 \\
        \midrule
        GRU-32 & MFB & 0.654 & 0.252 & 0.437 \\
        GRU-32 & W2V2-Fr-M-\textit{base} & \textbf{0.767} & \textbf{0.376} & \textbf{0.507} \\
        GRU-32 & XLSR-53-\textit{large} & 0.605 & 0.320 & 0.446 \\
        \midrule
        GRU-64 & MFB & 0.712 & 0.307 & 0.400 \\
        GRU-64 & W2V2-Fr-M-\textit{base} & \textbf{0.760} & \textbf{0.352} & \textbf{0.507} \\
        GRU-64 & XLSR-53-\textit{large} & 0.585 & 0.280 & 0.434 \\
        \bottomrule
    \end{tabular}}
\end{center}
\label{tab:EmoResults}
\end{table}
Best results are obtained by our W2V2-Fr-M-\textit{base} 
representation on valence, satisfaction and arousal, \cf Table \ref{tab:EmoResults}. With a simpler model, best scores are also achieved on both data sets with the Wav2Vec features, meaning that 
SSL representations are rich enough to be used with a simple regressor, even for low-quality speech signals (telephone conversations).

\section{Discussion}

After training our own SSL models for French, we evaluated them for four speech tasks (ASR, SLU, AST, and AER) using different architectures (shallow and deep architectures, end-to-end or not). The learnt SSL models are particularly beneficial for lower resource tasks (SLU, AST/TEDx, AER) or with simpler NN architectures (AER) but they sometimes fail providing a benefit compared to MFB or MFCCs (End-to-end ASR). Finetuning of SSL models could probably help bridging the gap remaining for some tasks, but we used SSL features `as they are' for this paper. 
Furthermore, efficient data augmentations techniques for Mel Filterbanks such as \textit{SpecAugment}
were disabled here to provide a comparison with SSL features, so we should highlight that we did not make the best of our Mel Filterbanks.   
All of these remarks and findings advocate for more exhaustive and reliable evaluations to assess the real impact of SSL for speech systems. We hope that decentralized projects such as {\nameproject} will contribute to this goal.


\section{Acknowledgements}
This work was performed using HPC resources from GENCI-IDRIS (Grant 2020-A0091012047). It was also partially supported by MIAI@Grenoble-Alpes (ANR-19-P3IA-0003).

\bibliographystyle{IEEEtran}

\bibliography{refs,refs2,refs3,refs4,main,anthology,laurent,mybib,datasets,shortened_bib}

\end{document}